\documentclass[letterpaper]{article} % DO NOT CHANGE THIS
\usepackage{aaai23}  % DO NOT CHANGE THIS
\usepackage{times}  % DO NOT CHANGE THIS
\usepackage{helvet}  % DO NOT CHANGE THIS
\usepackage{courier}  % DO NOT CHANGE THIS
\usepackage[hyphens]{url}  % DO NOT CHANGE THIS
\usepackage{graphicx} % DO NOT CHANGE THIS
\usepackage{natbib}  % DO NOT CHANGE THIS AND DO NOT ADD ANY OPTIONS TO IT
\usepackage{caption} % DO NOT CHANGE THIS AND DO NOT ADD ANY OPTIONS TO IT
\usepackage{color}
\usepackage{algorithm}
\usepackage{algorithmic}

\usepackage{newfloat}
\usepackage{listings}
\DeclareCaptionStyle{ruled}{labelfont=normalfont,labelsep=colon,strut=off} % DO NOT CHANGE THIS
\floatstyle{ruled}
\newfloat{listing}{tb}{lst}{}
\floatname{listing}{Listing}
\usepackage{amsmath,amssymb,amsfonts}

\title{Adversarial Policy Optimization in Deep Reinforcement Learning}
\author{
    Md Masudur Rahman,
    Yexiang Xue
}
\affiliations{
    Department of Computer Science, Purdue University \\
    West Lafayette, IN 47907, USA \\
    \{rahman64,yexiang\}@purdue.edu
}

\begin{document}

\maketitle

\begin{abstract}
The policy represented by the deep neural network can overfit the spurious features in observations, which hamper a reinforcement learning agent from learning effective policy. This issue becomes severe in high-dimensional state, where the agent struggles to learn a useful policy. Data augmentation can provide a performance boost to RL agents by mitigating the effect of overfitting. However, such data augmentation is a form of prior knowledge, and naively applying them in environments might worsen an agent's performance. In this paper, we propose a novel RL algorithm to mitigate the above issue and improve the efficiency of the learned policy. Our approach consists of a max-min game theoretic objective where a perturber network modifies the state to maximize the agent's probability of taking a different action while minimizing the distortion in the state. In contrast, the policy network updates its parameters to minimize the effect of perturbation while maximizing the expected future reward. Based on this objective, we propose a practical deep reinforcement learning algorithm, Adversarial Policy Optimization (APO). Our method is agnostic to the type of policy optimization, and thus data augmentation can be incorporated to harness the benefit. We evaluated our approaches on several DeepMind Control robotic environments with high-dimensional and noisy state settings. Empirical results demonstrate that our method APO consistently outperforms the state-of-the-art on-policy PPO agent. We further compare our method with state-of-the-art data augmentation, RAD, and regularization-based approach DRAC. Our agent APO shows better performance compared to these baselines.
\end{abstract}

\section{Introduction}
Reinforcement learning (RL) has achieved tremendous success in many applications, including simulated games (e.g., Atari, Chess, AlphaGo). 
The environments often provide observation which is a high-dimensional projection of the true state, complicating policy learning as the deep neural network model might mistakenly correlate reward with irrelevant information present in the observation. 
This scenario severely impaired the ability of the agent to learn and generalize even in a slightly modified environment \cite{Song2020Observational,cobbe2018quantifying,zhang2018study,machado2018revisiting,gamrian2019transfer}. 

\begin{figure}
    \centering
    \includegraphics[width=0.99\columnwidth]{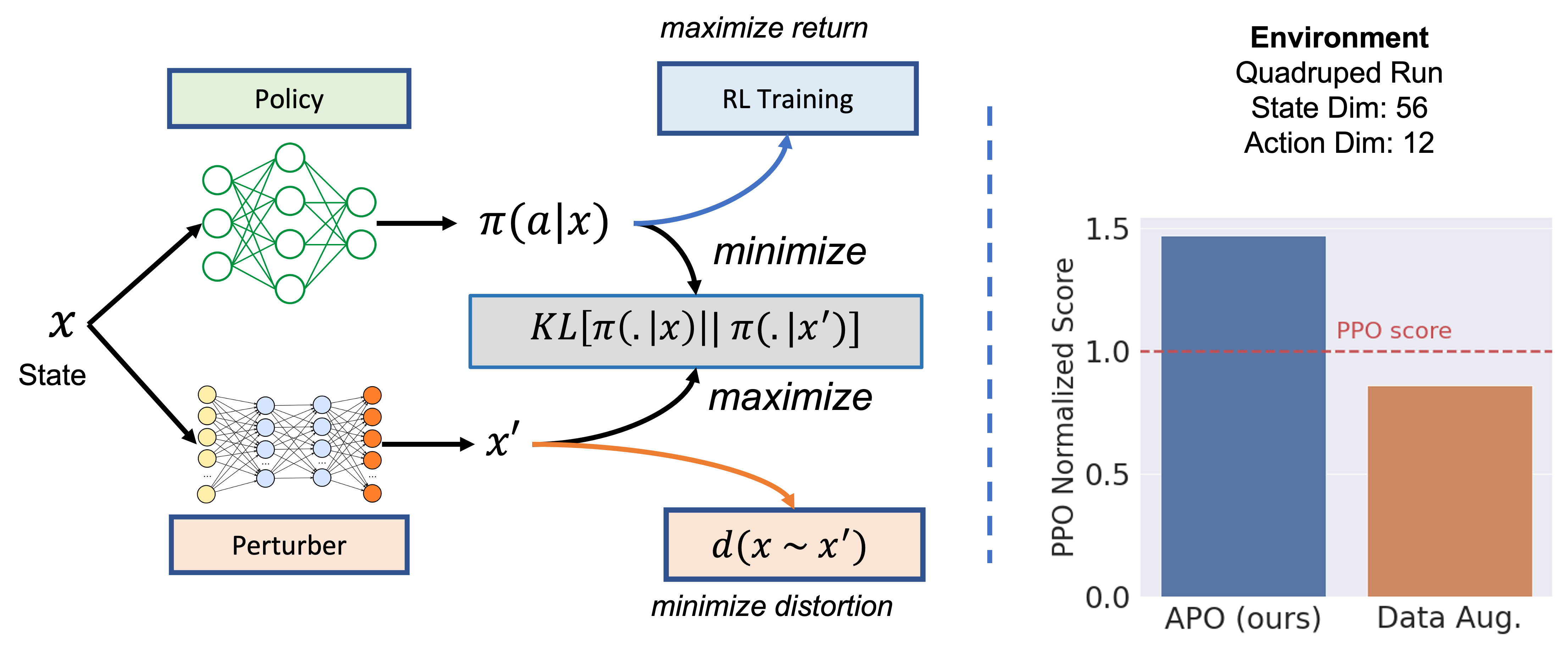}
    \caption{\small  [\textbf{Left}] Overview of our APO approach. The method consists of a max-min game theoretic objective where a perturber network maximizes the change in action probabilities of the policy measured in KL-divergence while minimizing the distortion of the observation. In contrast, the policy network updates its policy parameters to minimize the effect of perturbation while optimizing for the RL objective to maximize the expected future reward.
    [\textbf{Right}] Results on DeepMind Control Quadruped Run Environment. Our method APO \textbf{improves} the performance of the base algorithm by \textcolor{blue}{$47\%$} points, while data augmentation method (i.e., RAD \cite{laskin2020reinforcement}) \textbf{worsens} the performance by \textcolor{red}{$14\%$} points.}
    \label{fig:framework_result}
\end{figure}

In this paper, we focus on the challenge of high-dimensional state space where agents might overfit to a part of the state space in early training timestep and subsequently fails to recover, which results in poor performance. Moreover, the presence of noise in the state might make policy learning harder even in small state dimensions.
Data augmentation is a popular choice in improving sample efficiency in high-dimensional observation space \cite{cobbe2019quantifying,laskin2020reinforcement,raileanu2020automatic} which result in some empirical success. Data augmentation is a form of prior knowledge about the observation where the transformation is often assumed invariant to the task. However, improper handling of the observation transformation might lead to worse performance \cite{raileanu2020automatic}. Figure \ref{fig:framework_result} shows that on DeepMind Control \cite{tunyasuvunakool2020} Quadruped Run environment the data augmentation (i.e., RAD \cite{laskin2020reinforcement}) worsens the performance where PPO \cite{schulman2017proximal}-based method is used as the base policy. In contrast, our method achieves a $47\%$ points improved performance compared to the PPO. 

In this paper, we propose a novel algorithm that helps in learning robust policy in high-dimensional and noisy state spaces by an adversarial policy optimization approach. Figure \ref{fig:framework_result} shows an overview of our approach.
Our method consists of a max-min game theoretic objective where a perturber network modifies the observation to maximize the agent's probability of taking a different action while minimizing the distortion in the observation.
In contrast, the policy network (agent) updates its parameters to minimize the effect of perturbation while maximizing the expected future reward. 
Based on this objective, we propose a practical deep reinforcement learning algorithm, Adversarial Policy Optimization (APO). The adversarial max-min objective construction requires a policy ($\pi$) that maps the observation to action probabilities.

We empirically evaluate 6 environments with high-dimensional state space from DeepMind Control \cite{tunyasuvunakool2020}: Quadruped: Walk, Run, and Escape, and Dog: Walk, Run, and Fetch. In all these environments, APO outperforms PPO consistently. We further compare APO with state-of-the-arts pure data augmentation-based RAD \cite{laskin2020reinforcement} and DRAC \cite{raileanu2020automatic} which is a regularization-based method that leverages data augmentation.

We evaluate the robustness of APO in noisy states. We conduct experiments on 6 environments setting of Hopper, Walker, and Cheetah from DeepMind Control \cite{tunyasuvunakool2020}. These are relatively smaller state dimensions, and we augment each state with additional noise and increase the dimensions. In these settings, our method outperforms PPO consistently. We observed that the noise severely impacted PPO's performance while APO was was minimally impacted by it and stayed robust. 
We report the performance comparison in these noisy state setups with the baseline RAD and DRAC. Overall, our method shows better performance compared to these baselines. 

Furthermore, we compute the PPO normalized score where our method APO shows a performance improvement of $1.81$x and $7.95$x compared to the base PPO algorithm in high-dimensional and noisy state environments, respectively. Moreover, our method performs better than the data augmentation-based baselines RAD and DRAC in this measure.

In summary, our contributions are listed as follows:
\begin{itemize}
    \item We propose Adversarial Policy Optimization (APO), a deep reinforcement learning algorithm for high-dimensional and noisy states.
    \item We evaluate our approach on 10 DeepMind Control locomotion environment settings consisting of high-dimensional and noisy states. 
    \item Empirically, we observed that our agent APO outperforms PPO consistently in all settings, and our method achieves overall superior performance compared to state-of-the-art data augmentation-based approaches: RAD and DRAC.
\end{itemize}

\section{Preliminaries and Problem Settings} \label{sec:background}
\noindent\textbf{Markov Decision Process (MDP)}
is denoted by $\mathcal{M} =(\mathcal{S}, \mathcal{A}, \mathcal{P}, r)$. 
At every timestep $t$, from an state $s_t \in \mathcal{S}$ , the agent takes an action $a_t$ from a set of actions $\mathcal{A}$.
Agent receives a reward $r_t$ and the environment move to a new state 
$s_{t+1} \in \mathcal{S}$ based on the transition probability
$P(s_{t+1}\vert s_t, a_t)$. 

\noindent\textbf{Reinforcement Learning}
In reinforcement learning, the goal is to learn a policy $\pi \in \Pi$; that maximizes cumulative reward in an MDP, where $\Pi$ is the set of all possible policies. The policy is a mapping from states to actions, and an optimal policy $\pi ^*\in \Pi$ has the highest cumulative rewards. The agent interacts with an MDP environment (E) in discrete timesteps. At every timestep $t$, from an state $s_t$, the agent takes an action $a_t$ from a set of actions ($A$) based on the an observation ($o_t$). Agent receives a reward $r_t$ and the environment move to a new state $s_{t+1}$ based on a transition probability $Pr(s_{t+1}\vert s_t, a_t)$.

The term ``state" often refers to the actual state of the environment, and ``observation" refers to what the agent gets as input to carry out policy learning. Depending on various environment settings, these terms might be distinct in meaning. However, in our setup, they refer to the same entity. Thus, throughout the paper, we use state and observation alternatively.

\noindent\textbf{Policy Gradient} is a class of reinforcement learning algorithms where the objective is formulated to optimize the cumulative future return directly.
Proximal Policy Optimization (PPO) \cite{schulman2017proximal}, a type of policy gradient method which achieved tremendous success and is popularly used in many setups. The objective function of the PPO is as follows:
\begin{equation}\label{eq:policy_rl}
   \mathcal{L}_\pi = - \mathbb{E}_t[\frac{\pi_\theta(a_t \vert s_t)}{\pi_{\theta_{old}}(a_t \vert s_t)} A_t],
\end{equation}
where $\pi_\theta(a_t \vert s_t)$ is the probability of choosing action $a_t$ at state $s_t$ at timestep $t$ using the current policy. The current policy is parameterized by $\theta$ and the $\pi_{\theta_{old}}(a_t \vert s_t)$ refers to the probabilities using an old policy parameterized by previous parameter values $\theta_{old}$. $A_t$ is an advantage estimation which is the benefit of taking action $a_t$ at $s_t$ compared to some other measures such as the value function estimation at that state. The advantage estimation using a value function can be formulated as follows (equation \ref{eq:advantage}):
\begin{equation}\label{eq:advantage}
    A_t = -V(s_t) + r_t + \gamma r_{r+1} + ... + \gamma^{T-t+1} r_{T-1} + \gamma^{T-t} V(s_T),
\end{equation}
where $V(s)$ is a value function that gives the average future return under the current policy.

\noindent\textbf{Data augmentation} has been used to improve sample efficiency in various complex tasks \cite{laskin2020reinforcement,raileanu2020automatic}. 
This process aims to transform the observation while keeping its semantic meaning. Thus the changes only applied to contextual information changes, which are ideally irrelevant to the reward signal. This process depends on the specific task, and assumptions need to be made on which transformation is allowed. 

For vector-based states, various transformations (e.g., amplitude scaling) and in the case of image observation, various image manipulation can be used (e.g., random crop and color-jitter). In this paper, we focus on vector-based states. We use \textit{random amplitude scaling} proposed in RAD \cite{laskin2020reinforcement}. 
This method involves multiplying the observation with a number generated randomly between a range $\alpha$ to $\beta$. We used a range $\alpha=0.6$ to $\beta=1.2$ for all the agents that used data augmentation. This range is from the suggested range in \cite{laskin2020reinforcement} and, empirically, they work better in our setup.

In this paper, we evaluate and compare two data augmentation-based baselines RAD\cite{laskin2020reinforcement}  and DRAC \cite{raileanu2020automatic} . In RAD, the data augmentation is applied to the states before passing it to the policy network. In DRAC, the data augmentation is used to regularize the policy and value functions instead of passing directly to the policy.

\section{Adversarial Policy Optimization (APO)}
The method consists of a max-min game theoretic objective where a perturber network maximizes the policy's change in action probabilities while minimizing the distortion of the input observation. In contrast, the policy network (agent $\pi$) updates its policy parameters to minimize the perturbation effect while optimizing for the RL objective to maximize the expected future reward. Figure \ref{fig:framework_result} shows an overview of our framework.

Let $x$ be the observation the agent receives from the environment, and $x'$ is the perturbed output from the perturber network given the original observation, $x' = PerturberNetwork(x)$.
Intuitively, the perturber network tries to change $x$ to $x'$ to challenge the policy. In another word, the difference between $\pi(. \vert x)$ and $\pi(.\vert x')$ is maximized. 
If the difference measure in KL distance, the $KL[\pi(. \vert x)  \vert  \vert  \pi(. \vert x')]$ is maximized.
On the other hand, the policy network, $\pi$, tries to be robust to such distortion generated by the perturber network and thus update its parameter to minimize the $KL[\pi(. \vert x)  \vert  \vert  \pi(. \vert x')]$.

We now discuss the details of the perturber network and policy network.

\noindent\textbf{Perturber Network} It takes the observation $x$ as input and outputs the perturbed observation $x'$. The objective of this network is to minimize the amount of perturbation $(x \sim x')$ while maximally changing the policy behavior. Here ($x \sim x'$) denotes the measure of distortion from $x$ to $x'$. In our implementation we use $L_2$ norm, $ \vert\vert x - x' \vert\vert _2^2$ as the distortion measure.

\noindent Formally, given a policy $\pi_\theta$, the loss function, which the perturber network will minimize by updating parameter $\phi$ is defined as follows:
\begin{equation} \label{eq:perturb}
    \mathcal{L_\phi} =  \vert  \vert x_t-x_t' \vert  \vert _2^2 - KL(\pi_\theta(. \vert x_t), \pi_\theta(. \vert x_t')) 
\end{equation}

Here the first part of Equation \ref{eq:perturb} make sure the distortion is minimized, and the second part (i.e., KL part) participates in the adversarial optimization.

\noindent\textbf{Policy Network}
The goal of the policy network is to learn a robust policy. The adversarial optimization occurs on top of the RL objective of the policy, $\mathcal{L_\pi}$. Thus the overall optimization happens by minimizing the loss in Equation \ref{eq:policy}. 

\begin{equation}\label{eq:policy}
    \mathcal{L_\theta} =  \mathcal{L}_\pi + KL[\pi_\theta(. \vert x_t), \pi_\theta(. \vert x_t')] 
\end{equation}

The first part $\mathcal{L_\pi}$ is the RL objective during training. The first part, $\mathcal{L_\pi}$ (as in Equation \ref{eq:policy_rl}),  ensures the learned policy achieves maximum cumulative reward. On the other hand, the second part of the Equation \ref{eq:policy} participates in the adversarial optimization.

\begin{algorithm}
\caption{Adversarial Policy Optimization (APO)}
\label{algo-apo}
\begin{algorithmic}[1]
    \STATE Initialize parameter vectors $\theta$ for policy network, and $\phi$ for perturber network
    % \State Return $\mathcal{C}$
    \FOR {each iteration}
        \FOR{each environment step} %\Comment{data collection}
            \STATE $a_t \sim \pi_\theta(a_t \vert x_t)$ %\Comment{using current policy}
            \STATE $x_{t+1} \sim P(x_{t+1} \vert x_t, a_t)$ %\Comment{from environment}
            \STATE $r_t \sim R(x_t, a_t)$ %\Comment{from environment}
            \STATE $\mathcal{D} \leftarrow{} \mathcal{D} \cup \{(x_t, a_t, r_t, x_{t+1})\} $ %\Comment{store data in buffer}
        \ENDFOR
        
        \FOR{each observation $x_t$ in $\mathcal{D}$}
        % \Comment{optimize for $\mathcal{L}_\theta$, and $\mathcal{L}_\phi$} // optimize for $\mathcal{L}_\theta$, and $\mathcal{L}_\phi$
        \STATE $x_t' \leftarrow{} PerturberNetwork(x_t)$ 
        % \Comment{generate perturbed observation} //generate perturbed observation
        \STATE Compute $\mathcal{L}_\pi$ from data $\mathcal{D}$ using base RL algorithm (i.e., PPO)
        % \Comment{RL policy specific loss}// RL policy specific loss
        \STATE Compute $\mathcal{L}_\theta$ using Equation \ref{eq:policy} 
        % \Comment{policy network loss} // policy network loss
        \STATE Update $\theta$ using backpropagation.
        
        \STATE Compute $\mathcal{L}_\phi$ using Equation \ref{eq:perturb} 
        % \COMMENT{perturb network loss} //perturb network loss
        \STATE Update $\phi$ using backpropagation.
        \ENDFOR
    \ENDFOR
\end{algorithmic}
\end{algorithm}

Finally, we develop a practical reinforcement learning algorithm using the above formulation. Details of the algorithm are given in Algorithm \ref{algo-apo}.
At each timestep $t$, the two losses $\mathcal{L}_\theta$, and $\mathcal{L}_\phi$ are minimized alternatively by updating parameters $\theta$, and $\phi$ respectively, using back-propagation.

Note that, Algorithm \ref{algo-apo} shows general steps to incorporate our method into base RL algorithms. The adversarial optimization requires accessing a policy mapping function that maps state-to-action distribution. We further show the implementation, assuming the policy network is implemented using a neural network. APO adds a perturber loss objective that uses the policy ($\pi$) to compute the KL-term in equation \ref{eq:perturb}. On the other hand, the policy loss or actor loss ( $\pi$) is regularized by the KL-component as in equation \ref{eq:policy}. Our adversarial objective is complementary to other methods such as data augmentation (e.g., RAD). Thus, this method can be incorporated to harness its benefit of it. In that case, an additional step would be to change the data observation $x_t$ before passing it to the Perturber and Policy networks. Unless otherwise specified in this paper, we use data augmentation on the observation before passing it to the Perturber and Policy networks. We also compare our method with these data augmentation-based techniques and show the empirical advantage of our APO algorithms. In this paper, we demonstrate the usefulness of our method in the policy gradient-based algorithms (e.g., PPO). 

\section{Experiments}
\begin{figure}
    \centering
    \includegraphics[width=0.99\columnwidth]{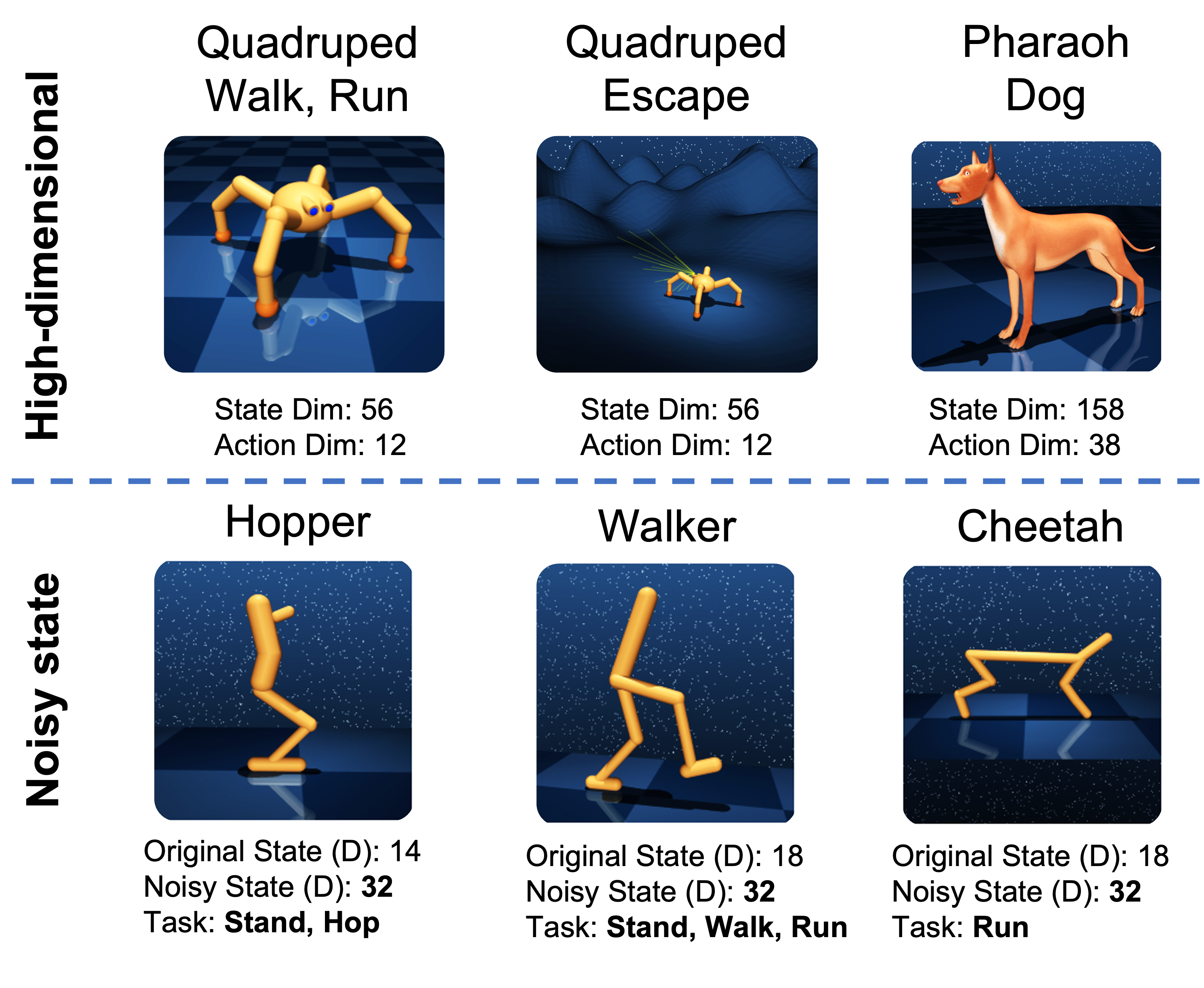}
    \caption{[\textbf{Top}] DeepMind Control Environments \cite{tunyasuvunakool2020} with high-dimensional observation.  
    These are challenging tasks due to high-dimensional state, action spaces, and complex behavior. [\textbf{Bottom}] DeepMind Control Environments with the noisy state. For each environment, random noise is appended with the original states at each timestep which extends the state dimension to 32. Thus the tasks become challenging, and the agent must avoid the effect of noise to achieve optimal behavior. 
    }
    \label{fig:dmc_env}
\end{figure}
In this section, we discuss our experiment setup and analyze the results.

\subsection{Settings} 
Our implementation of PPO is based on the \cite{shengyi2022the37implementation,huang2021cleanrl} where many important details are implemented which are curated from the research in recent years on policy gradient (e.g., Normalization of Advantage, Orthogonal Initialization, GAE). We refer reader to \cite{shengyi2022the37implementation} for further references. Note that this implementation consists of all these improvements, and we would like to acknowledge all the past research in addition to the original PPO paper \cite{schulman2017proximal}. We still call this algorithm PPO to be consistent with the existing literature. Our agent APO and baselines RAD and DRAC use the same implementation for a fair comparison.

We use the same hyperparameter for the base PPO algorithms for all experiments and agents. These hyperparameters are based on the best practices reported in the PPO implementation of continuous action spaces \cite{shengyi2022the37implementation,huang2021cleanrl}. 
\begin{table}
\caption{Hyperparameters for the experiments.} 
\label{tab:ppo-hyp} 
\begin{center}
 % \scriptsize
\begin{tabular}{c|c}
\hline
 \textbf{Description} & \textbf{Value}  \\
\hline
Number of rollout steps  & $2048$  \\
\hline
Learning rate & $3e-4$ \\
\hline
Discount factor gamma & $0.99$\\
\hline
Lambda for the GAE  & $0.95$ \\
\hline
Number of mini-batches  & $32$ \\
\hline
Epochs to update the policy & $10$ \\
\hline
Advantages normalization & True \\
\hline
Surrogate clipping coefficient & $0.2$\\
\hline
Clip value loss & Yes\\
\hline
Value loss coefficient &$0.5$\\
\hline
\end{tabular}
\end{center}
\end{table}
We report the common hyperparameters in Table \ref{tab:ppo-hyp}. 

\subsection{Results in high-dimensional observation}
\begin{figure*} %[tbp]
  \centering
  \begin{minipage}[b]{0.33\textwidth}
    \includegraphics[width=0.99\textwidth]{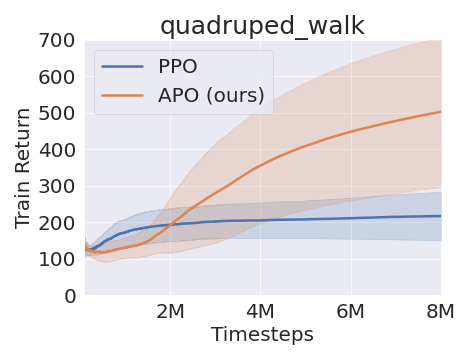}
  \end{minipage}
  \hfill
  \begin{minipage}[b]{0.33\textwidth}
    \includegraphics[width=0.99\textwidth]{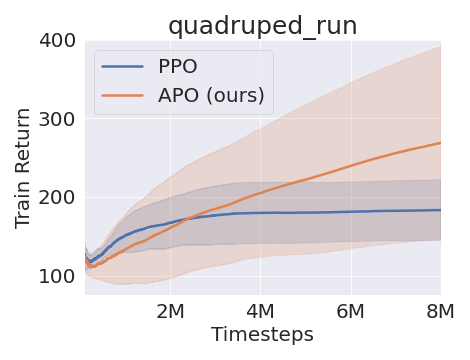}
  \end{minipage}
  \hfill
  \begin{minipage}[b]{0.33\textwidth}
    \includegraphics[width=0.99\textwidth]{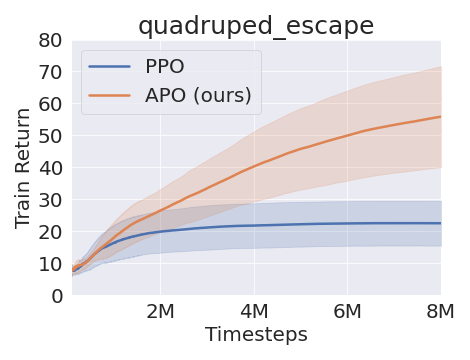}
  \end{minipage}
  \hfill
  \begin{minipage}[b]{0.33\textwidth}
    \includegraphics[width=0.99\textwidth]{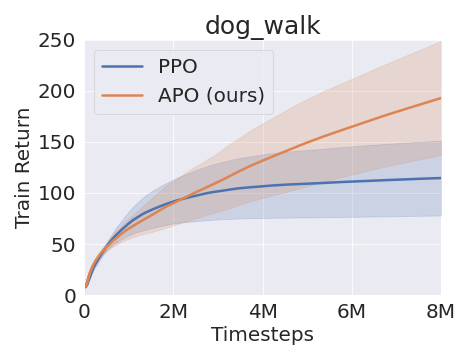}
  \end{minipage}
  \hfill
  \begin{minipage}[b]{0.33\textwidth}
    \includegraphics[width=0.99\textwidth]{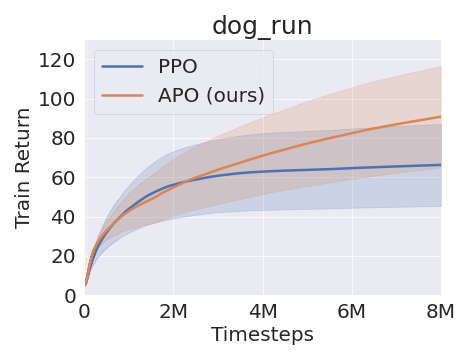}
  \end{minipage}
  \hfill
  \begin{minipage}[b]{0.33\textwidth}
    \includegraphics[width=0.99\textwidth]{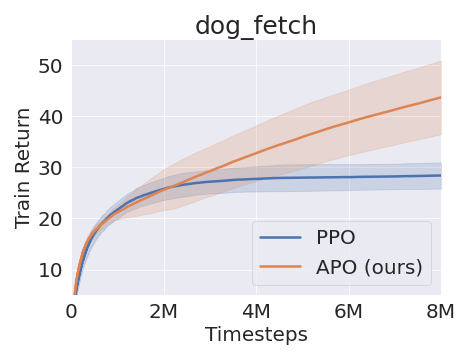}
  \end{minipage}
  \hfill
  \caption{Results on environments with high-dimensional states. Our agent APO outperforms the PPO algorithms in all the environments. In most scenarios, the PPO agent learns up to 2M timesteps and then fails to make reasonable progress. In contrast, our method APO consistently keeps improving the performance as the timestep increases. The mean and standard deviations are over 10 seed runs.
  }
  
  \label{fig:apo_ppo_high_dim_return}
\end{figure*}
To demonstrate the effectiveness of our method, we conducted experiments on the challenging simulated robotic environments, shown in Figure \ref{fig:dmc_env}, Top. These environments are challenging as both state and action spaces are large. In these setups, the agent's policy network might overfit the part of the state and learn a sub-optimal policy.

Figure \ref{fig:apo_ppo_high_dim_return} shows result comparison of APO with PPO in high-dimensional state spaces setup. We see that our agent APO achieves reasonable performance and consistently makes progress as the timestep increases. In all these environments, APO eventually outperforms the PPO agents. PPO agent fails to make any reasonable progress after about 2M timesteps in most of the scenarios. These results show the advantage of our adversarial training in high-dimensional states.

We further compare the performance of our method APO with the data augmentation-based approach RAD and DRAC. Table \ref{tab:comparison_results_high_dim} shows the performance of each agents after training the agents for 8M timesteps.
We see that our method APO performs better than RAD in 4 out of 6 environments. APO performs better than DRAC in 5 out of 6 environment. In many scenarios, data augmentation techniques RAD and DRAC improve the performance of base PPO algorithms. These results show the importance of data augmentation in such challenging high-dimensional tasks. Overall, our adversarial policy training achieves the best performance, indicating our method's importance in such setups. 
An essential aspect of the agent evaluation is that it has to maintain the base algorithm performance and not make the performance worse. However, we observe that in the baseline data augmentation approach, RAD shows worsening performance in the Quadruped Run environment (Figure \ref{fig:framework_result}, and Table \ref{tab:comparison_results_high_dim}). In contrast, our method APO consistently outperforms the base PPO agents and does not worsen the performance in the evaluated environments. These results show the consistency of our proposed method in learning policy in complex scenarios.

\subsection{Results in noisy state environments}
We further test the robustness of our method APO by evaluating it for tasks with noisy states. We use 6 DeepMind control environments for the experiments as shown in Figure \ref{fig:dmc_env}, Bottom. 
These are relatively smaller state spaces where Hopper has state dimension 14, and Walker and Cheetah have state dimension 18. We modified the state dimension by appending noise into the state at each timestep of the environment. The goal here is to keep the state's information the same while adding some additional dimension with noise. This modification injects noise and increases the state's dimensionality; thus the environments become more challenging than the original tasks. 

In particular, we increase the dimension of all the environments to 32 by appending additional 18 (Hopper) and 14 (Walker and Cheetah) dimensional noise vector. To compute each noise element, we consider the original state element as the mean of a Gaussian distribution with a standard deviation of $1.0$. We draw a sample from this distribution to generate a noise element. Thus, we add the noise sequentially from the original state elements and keep adding unless the total dimension reaches 32. Note that each state contains the unchanged original state information; thus, a robust agent would still be able to learn an effective policy.

\begin{table*}
\caption{Comparison on high-dimensional state environments. Our agent APO outperforms PPO by a large margin and achieves better performance than baselines: RAD and DRAC. The results are the final performance after training the agents for 8M environment timesteps. The mean and standard deviations are over 10 seed runs.
} 
\label{tab:comparison_results_high_dim} 
\begin{center}
 % \scriptsize
\begin{tabular}{c|c|c|c|c}
\hline
\textbf{Env} & \textbf{PPO} & \textbf{RAD-PPO} & \textbf{DRAC-PPO} & \textbf{APO (ours)} \\
\hline \hline
quadruped walk & 216.67 \small{$\pm$66.56} & 395.07 \small{$\pm$124.91} & 374.78 \small{$\pm$197.98} & \textbf{502.16} \small{$\pm$205.23}\\
\hline
quadruped run & 183.14 \small{$\pm$38.48} & 158.01 \small{$\pm$26.84} & 247.02 \small{$\pm$83.23} & \textbf{268.63} \small{$\pm$122.19}\\
\hline
quadruped escape & 22.51 \small{$\pm$7.0} & 53.94 \small{$\pm$22.1} & 52.56 \small{$\pm$21.86} & \textbf{55.85} \small{$\pm$15.79}\\
\hline
dog walk & 114.55 \small{$\pm$36.63} & \textbf{211.89} \small{$\pm$33.25} & 205.68 \small{$\pm$42.89} & 192.76 \small{$\pm$55.68}\\
\hline
dog run & 66.25 \small{$\pm$20.86} & 85.35 \small{$\pm$29.08} & 88.9 \small{$\pm$8.33} & \textbf{90.77} \small{$\pm$25.8}\\
\hline
dog fetch & 28.4 \small{$\pm$2.57} & \textbf{46.26} \small{$\pm$9.21} & 42.98 \small{$\pm$7.35} & 43.68 \small{$\pm$7.21}\\
\hline
\end{tabular}
\end{center}
\end{table*}

\begin{figure*}[!tbp]
  \centering
  \begin{minipage}[b]{0.33\textwidth}
    \includegraphics[width=0.99\textwidth]{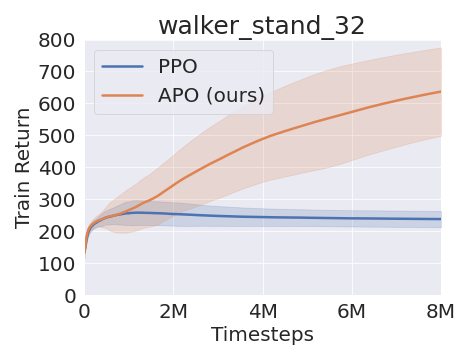}
  \end{minipage}
  \hfill
  \begin{minipage}[b]{0.33\textwidth}
    \includegraphics[width=0.99\textwidth]{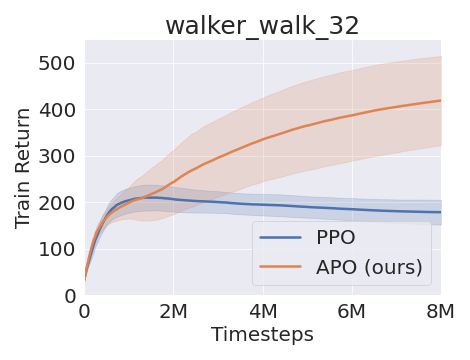}
  \end{minipage}
  \hfill
  \begin{minipage}[b]{0.33\textwidth}
    \includegraphics[width=0.99\textwidth]{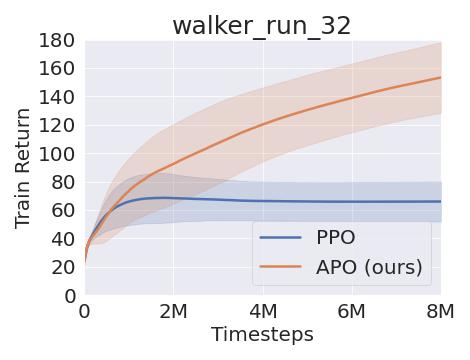}
  \end{minipage}
  \hfill
  \begin{minipage}[b]{0.33\textwidth}
    \includegraphics[width=0.99\textwidth]{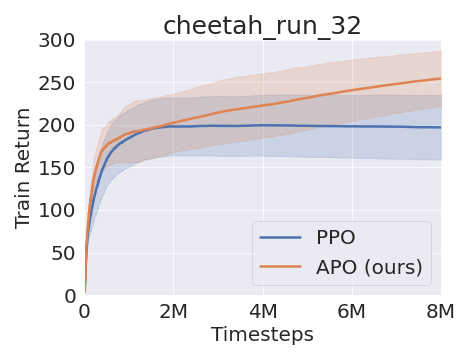}
  \end{minipage}
  \hfill
  \begin{minipage}[b]{0.33\textwidth}
    \includegraphics[width=0.99\textwidth]{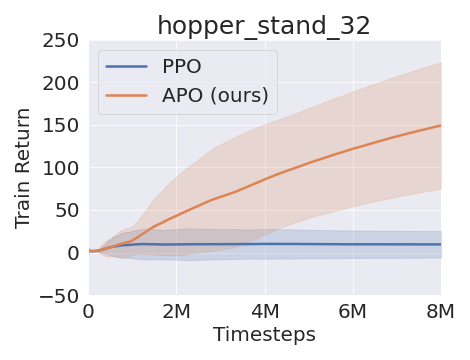}
  \end{minipage}
  \hfill
  \begin{minipage}[b]{0.33\textwidth}
    \includegraphics[width=0.99\textwidth]{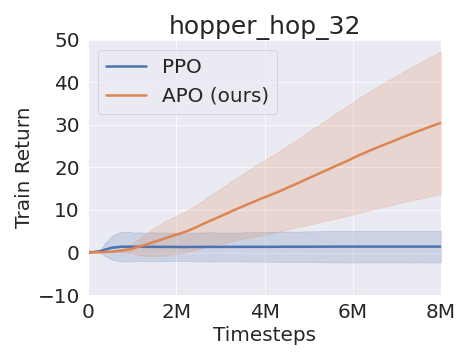}
  \end{minipage}
  \hfill
  \caption{Results on environments with noisy states. Our agent APO outperforms the PPO in all the environments. In most scenarios, the PPO agent fails to make reasonable progress due to the noisy and extended state dimensions. In contrast, our method APO consistently keeps improving the performance as the timestep increases showing robustness against noise and extended dimension. The mean and standard deviations are over 10 seed runs.
  }
  \label{fig:apo_ppo_hdo_return}
\end{figure*}

Figure \ref{fig:apo_ppo_hdo_return} shows the results comparison of our APO with PPO. We see that in all six environments, APO outperforms the PPO agents. In most scenario, the PPO agent stops learning after about 1M timestep and stay sub-optimal. PPO fails to learn any reasonable return in Hopper (Stand and Hop) environments. In contrast, our APO shows robustness against the noise and extended dimensionality, consistently outperforms the PPO, and keeps learning as the timestep increases. These results show the advantage of our method in dealing with this challenging setup.

\begin{table*}
\caption{Comparison in noisy state environments. Our agent APO outperforms PPO by a large margin and achieves better performance than data augmentation-based baselines (RAD, DRAC). The mean and standard deviations are over 10 seed runs.
} 
\label{tab:comparison_results_dmc_hdo} 
\begin{center}
 % \scriptsize
\begin{tabular}{c|c|c|c|c}
\hline
\textbf{Env} & \textbf{PPO} & \textbf{RAD-PPO} & \textbf{DRAC-PPO} & \textbf{APO (ours)} \\
\hline \hline
walker stand 32 & 238.22 \small{$\pm$25.43} & 576.84 \small{$\pm$181.31} & 278.32 \small{$\pm$74.3} & \textbf{637.05} \small{$\pm$138.27}\\
\hline
walker walk 32 & 154.22 \small{$\pm$27.77} & 522.07 \small{$\pm$95.13} & 475.44 \small{$\pm$135.33} & \textbf{585.26} \small{$\pm$68.77}\\
\hline
walker run 32 & 65.95 \small{$\pm$14.07} & 142.12 \small{$\pm$36.25} & 99.46 \small{$\pm$16.83} & \textbf{153.21} \small{$\pm$24.89}\\
\hline
cheetah run 32 & 193.71 \small{$\pm$30.13} & 312.47 \small{$\pm$40.67} & \textbf{328.92} \small{$\pm$36.0} & 324.34 \small{$\pm$28.72}\\
\hline
hopper stand 32 & 9.61 \small{$\pm$15.53} & 131.89 \small{$\pm$92.86} & 47.63 \small{$\pm$39.58} & \textbf{149.07} \small{$\pm$74.14}\\
\hline
hopper hop 32 & 1.4 \small{$\pm$3.71} & 20.14 \small{$\pm$16.48} & 9.69 \small{$\pm$12.18} & \textbf{30.42} \small{$\pm$16.73}\\
\hline
\end{tabular}
\end{center}
\end{table*}
We further compare our method with data augmentation-based baseline RAD and DRAC. Table \ref{tab:comparison_results_dmc_hdo} shows the results after training the agent for 8M timesteps. We see that our agent APO outperforms PPO by a large margin in all the environments. APO outperforms other data augmentation baselines in all the environments except the cheetah run, where DRAC slightly performs better than APO. Overall, the data augmentation baseline RAD and DRAC perform better than the PPO. In most of the scenarios, RAD performs better compared to DRAC. Overall, our method APO shows consistently better performance than all the baselines. These results demonstrate that our method is robust to these noise settings where the base agent performs poorly.

\noindent\textbf{Comparison on Aggregate Score}
Our evaluated environments are diverse, thus their episodic return value varies. Therefore, we further compute the aggregate scores to understand the overall improvement of our method APO over PPO and other baselines. We compute the PPO normalized score on 6 high-dimensional and 6 noisy state environments. For each type of environment, we normalize the mean performance of each agent by the mean score of the PPO; thus, the PPO's score is $1.0$. For each environment type (high-dim and noisy state), we report the average normalized score.
Table \ref{tab:comparison_results_ppo_normalzied} shows the results. In this measure, our APO achieves an $81\%$ point improvement over base agent PPO in high-dimensional environments. Our method also outperforms the other two baselines, RAD and DRAC, where they achieve $64\%$ and $68\%$ point improvement over PPO, respectively.

\begin{table*}
\caption{Comparison of PPO Normalized Score. Our method APO improves over PPO and performs better than other baselines. The mean return is first normalized using the mean return of the PPO, and then the average normalized scores are reported for each environment type.
} 
\label{tab:comparison_results_ppo_normalzied} 
\begin{center}
 % \scriptsize
\begin{tabular}{c|c|c|c|c}
\hline
\textbf{Env Type} & \textbf{PPO} & \textbf{RAD-PPO} & \textbf{DRAC-PPO} & \textbf{APO (ours)} \\
\hline \hline
High-Dim & 1.0 & 1.64 & 1.68 & \textbf{1.81} \\
\hline 
Noisy State & 1.0 & 6.28 & 3.22 & \textbf{7.95} \\
\hline
\end{tabular}
\end{center}
\end{table*}
Our method APO achieves a $7.95$x performance improvement in noisy state environments compared to the base PPO algorithm. Moreover, APO performs better compared to RAD and DRAC, where they show an improvement of $6.8$x and $3.22$x compared to PPO, respectively. The overall performance gain compared to PPO shows that the noisy state environment setup can still be solvable, and reasonable progress can be made by potentially mitigating the effect of the noise. However, the standard PPO baseline fails to perform consistently and often performs sub-optimally in these tested environments.
Overall, these results demonstrate the robustness of APO in the presence of noise in environments compared to other methods evaluated.
\section{Related Work} 
In recent time, many approaches have been proposed to address the challenges of reducing agent's overfitting to observation \citep{cobbe2019quantifying,laskin2020reinforcement,raileanu2020automatic, laskin2020curl}. Other approaches to the issue are
random noise injection \citep{igl2019generalization}, network randomization \citep{osband2018randomized,burda2018exploration,Lee2020Network}, and regularization \citep{cobbe2019quantifying,igl2019generalization,wang2020improving}.
The common theme of these approaches is to increase diversity in the training data so that the learned policy potentially mitigates overfitting. In contrast, our method learns an adversarial perturbation, which eventually guides the agent to avoid overfitting to high-dimensional and noisy states and thus eventually improves the performance.
The adversarial method has been used for imitation, and inverse reinforcement learning context 
\cite{barde2020adversarial,ho2016generative,henderson2018optiongan}. The general idea is to use the generator-discriminator setup to learn policy often from an expert demonstration. In contrast, our method uses an adversarial approach to learn policy in a reinforcement learning setup.
The adversarial attack has been explored \cite{zhang2020robust,Gleave2020Adversarial,lin2017tactics} in the context of reinforcement learning. The general setup is to intentionally have an adversary modify the state or hamper the agent's behavior. In contrast to this approach, our method uses the adversary during policy learning to help learn a robust policy.
Adversarial setup has been used to learn robust policy learning with varying environment dynamics \cite{kuang2022learning,pinto2017robust}, often assuming access to environment dynamics (e.g., mass, friction) \cite{pinto2017robust}.
In contrast, our method targets learning a robust policy on high-dimensional and noisy state environments that does not require access to environment dynamics; instead, our perturber network learns to generate perturbation on the state.

Auto-encoder based models with reconstruction loss have been proposed to learn latent state \citep{lange2012autonomous,lange2010deep,hafner2019learning}. However, the reconstruction loss does not guarantee state representation learning as some noise in the input might be co-related and thus important for the reconstruction while not necessarily important for the reward. 
In recent time, state learning approaches \citep{higgins2017darla,agarwal2021pse,zhang2020learning} have been proposed. The sequential structure in RL \citep{agarwal2021pse} and behavioral similarity between states \citep{zhang2020learning} have been leveraged to improve RL robustness. \cite{zhang2020learning} proposed to learn invariant state based on the assumption that the bisimulation metrics can help to learn such learn invariant representation even without reconstruction loss. 

In contrast to the above methods, we propose an adversarial max-min optimization where a perturber network compete with the policy network eventually helps the policy to be robust to noise along with high-dimension. We demonstrated the usefulness of our method on vector-based state space empirically and showed that our method helps to improve the robustness of the baseline algorithm in various settings.
\section{Conclusion}
We propose a novel RL algorithm to mitigate the issue when learning from high-dimensional and noisy states. Our method APO consists of a max-min game theoretic objective where a perturber network modifies the state to maximize the agent's probability of taking a different action. Furthermore, the policy network updates its parameters to minimize the effect of perturbation. We evaluated our approaches in several simulated robotic environments with high-dimensional and noisy settings. Empirical results demonstrate that our method APO consistently outperforms state-of-the-art on-policy  PPO and performs better than data augmentation-based baselines: RAD and DRAC.

% \section{Acknowledgments}
% 
\bibliography{main_ref}
\end{document}